# Riemannian kernel based Nyström method for approximate infinite-dimensional covariance descriptors with application to image set classification

Kai-Xuan Chen[1], Xiao-Jun Wu[1,*], Rui Wang[1], Josef Kittler[2]
[1] School of IoT Engineering, Jiangnan University, 214122, Wuxi, China
[2] Center for Vision, Speech and Signal Processing(CVSSP), University of Surry, GU2 7XH, Guildford, UK
{kaixuan_chen_jnu, xiaojun_wu_jnu}@163.com, RunningWang@outlook.com, j.kittler@surrey.ac.uk

*Abstract*—In the domain of pattern recognition, using the CovDs (Covariance Descriptors) to represent data and taking the metrics of the resulting Riemannian manifold into account have been widely adopted for the task of image set classification. Recently, it has been proven that infinite-dimensional CovDs are more discriminative than their low-dimensional counterparts. However, the form of infinite-dimensional CovDs is implicit and the computational load is high. We propose a novel framework for representing image sets by approximating infinite-dimensional CovDs in the paradigm of the Nyström method based on a Riemannian kernel. We start by modeling the images via CovDs, which lie on the Riemannian manifold spanned by SPD (Symmetric Positive Definite) matrices. We then extend the Nyström method to the SPD manifold and obtain the approximations of CovDs in RKHS (Reproducing Kernel Hilbert Space). Finally, we approximate infinite-dimensional CovDs via these approximations. Empirically, we apply our framework to the task of image set classification. The experimental results obtained on three benchmark datasets show that our proposed approximate infinite-dimensional CovDs outperform the original CovDs. (Source Code: https://github.com/Kai-Xuan/AidCovDs)

## I. Introduction

Image set classification has received a lot of attention in artificial intelligence and pattern recognition [1,2,3,4,5,6,7,8]. In the domain of image set classification, each set contains a number of images, possibly acquired in different environments, which have the same label. To this end, image sets can offer more discriminative and robust information than a single-shot image [9,10,11,12,13]. The commonly used representations of image sets include, linear subspaces [1], CovDs(covariance descriptors) [7,8,15], Gaussian mixture model [14], among which, CovDs defined by second-order statistics of image features have been widely applied in object detection [4], gesture classification [5], and virus recognition [13].

The CovDs are in the form of SPD(Symmetric Positive Definite) matrices which lie on a non-linear manifold known as the SPD manifold [7,8]. The space spanned by the SPD matrices is not a vector space because it does not satisfy the scalar multiplication axiom. For example, multiplying an SPD matrix by a negative scalar will result in a negative definite matrix instead of SPD matrix [11]. As a consequence, Euclidean metric cannot be used to analyze SPD matrices. Instead, Riemannian metrics have been shown to be a better tool for operating on SPD matrices. To this end, the SPD manifold, which forms the interior of a convex cone in the Euclidean space, is one kind of Riemannian manifold. A variety of Riemannian metrics of SPD manifold have been proposed. In particular, the AIRM(Affine Invariant Riemannian Metric) [2,8], which measures the geodesic distance on SPD manifold, and has the property of affine invariance, is the most popular. The Stein divergence and Jeffrey divergence [2,10] are efficient metrics to measure geodesic distance between two SPD matrices, which are derived from Bregman divergence for some special seed functions. The LEM(Log-Euclidean Metric) [7,8] views the SPD matrices on the Lie group. It measures the similarity between two SPD matrices through computing the distance in the SPD matrix logarithm domain, which is a flat surface at the point of identity matrix.

Recently, infinite-dimensional CovDs have become increasingly popular because of their robustness for the task of recognition as argued in [16,17,18]. In these three papers, the features of the samples are mapped to the RKHS(Reproducing Kernel Hilbert Space) through a kernel function, i.e., $\varphi: R^d \to \mathcal{H}$. Note, the infinite-dimensional CovDs in RKHS do not have an explicit representation form. To tackle this problem, in [16], the authors extend the Bregman divergence to analyze the infinite-dimensional CovDs in RKHS via the kernel trick. Similarly, in [17] and [18], the LEM and AIRM were extended to the infinite-dimensional setting. However, the dimension of features in RKHS is infinite and the number of independent observations is finite, which inevitably results in the rank of infinite-dimensional CovDs being deficient. To overcome this drawback, in [13] and [19], the authors propose to approximate infinite-dimensional CovDs in RKHS with an explicit mapping. The resulting CovDs are effective for image classification by virtue of approximating explicit forms of infinite-dimensional CovDs in RKHS.

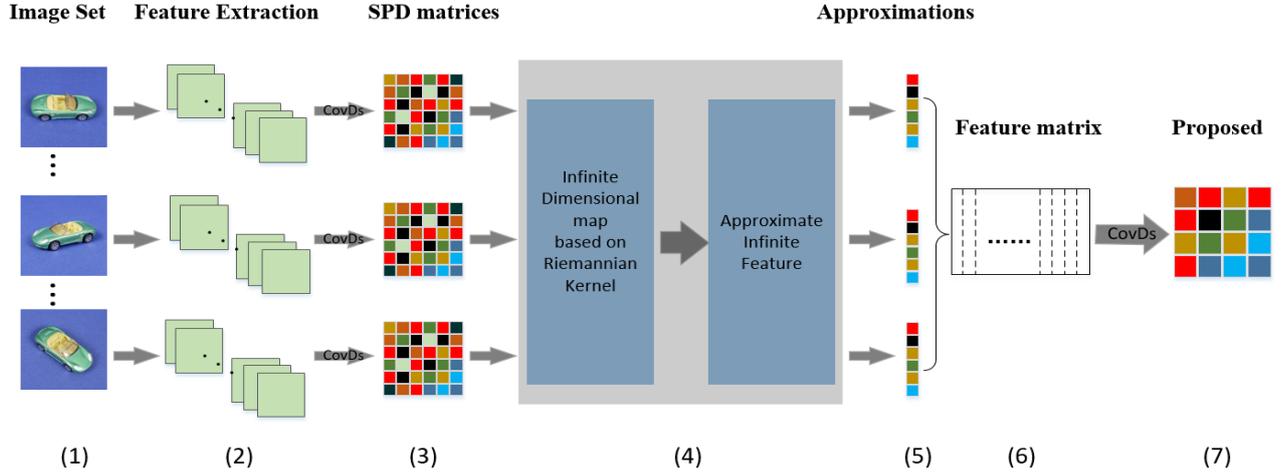

Fig.1. The flow chart of our proposed framework. Given an image set (Step 1). Extract the features of images (Step 2). Use the robust CovDs to represent the images in image set (Step 3). Map the CovDs into RKHS via Riemannian kernel and compute associated parameters for Nyström method (Step 4). Obtain the approximations of CovDs in RKHS (Step 5). Combine the approximations to form a feature matrix and compute its covariance matrix to represent the given image set (Step 6 and Step 7).

In this paper, we propose a novel framework to approximate infinite-dimensional CovDs, for use as data descriptors of image sets. Different from [13], we approximate infinite-dimensional CovDs by the Nyström method based on a Riemannian kernel and tackle the task of image set classification instead of image classification. In the traditional CovDs for image sets, the images need to be resized and vectorized for CovDs, which will result in the loss of information. Compared with the traditional data descriptors for image sets, the representation obtained by our proposed framework has the following three advantages: (1) we use the CovDs to represent the images without resizing and vectorizing them, which leads to more robust data descriptors for each single image in the sets. (2) our proposed framework gives an explicit form of CovDs to approximate infinite-dimensional CovDs in RKHS, which is a very competitive descriptor for image sets. (3) For the image sets, the dimensionality of traditional CovDs is usually $400 \times 400$ by resizing the images to $20 \times 20$. In this paper, the dimensionality of CovDs obtained by the proposed framework is $40 \times 40$ or $80 \times 80$, which are far lower than $400 \times 400$. Fig.1 gives the flow chart of the proposed framework. For a given image set (Fig.1 Step 1), we firstly extract features of images via Gabor filter [10,11,13] or SIFT [22] (Fig.1 Step 2) and use the CovDs of these features to represent the images (Fig.1 Step 3). Secondly, we map the CovDs of images into RKHS via Riemannian kernel and compute the associated parameters for the Nyström method (Fig.1 Step 4), and use them to approximate CovDs in RKHS (Fig.1 Step 5). Finally, we combine the approximations of the images to form a feature matrix and compute its covariance matrix to represent the given image set (Fig.1 Step 6 and Step 7).

The rest of this paper is organized as follows: In Section II, we give a brief overview of CovDs for image sets and infinite-dimensional CovDs, and introduce some classical related Riemannian metrics of SPD manifold. In Section III, we present the model of Nyström method and our proposed framework, and introduce some SPD manifold-based classification algorithms which are used in the experiments of this paper. In Section IV, we present the experimental results in terms of average accuracies and standard deviations. To this end, the results of the experiments show that the descriptors of our proposed framework offer more discriminative information than the traditional ones for the task of image set classification. In Section V, we express our conclusions and suggest future work.

## II. RELATED WORK

In this section, we give an overview of the traditional CovDs and infinite-dimensional CovDs for image sets, and introduce some classical Riemannian metrics of SPD manifold. In this paper, we will adopt the following notations: $S_n^+$ is the space spanned by real $n \times n$ SPD matrices. $T_P S_n^+$ is the tangent space at the point $P \in S_n^+$, where the dimensionality of each points is $n \times n$. $S_n$ is the flat surface spanned by real $n \times n$ symmetric matrices, which is the tangent space at the point of identity matrix $I_n \in R^{n \times n}$.

### A. Traditional CovDs and Infinite-Dimensional CovDs

Given an image set with $n$ images, $S = [s_1, s_2 \ldots, s_N]$, where $s_i \in R^D$ is obtained by resizing and vectorizing the $i$-th image sample of image set. Let $C$ be the covariance matrix [2,7,8,13] computed from the raw intensity of resized samples in the image sets:

$$C = \frac{1}{n} \sum_{i=1}^{N} (s_i - \bar{s})(s_i - \bar{s})^T = S J_N J_N^T S^T \quad (1)$$

where $\bar{s} = \frac{1}{N} \sum_{i=1}^{N} s_i$ is the mean vector of image samples in the set $S$, and $C$ is a $D \times D$ covariance matrix. $J_N = N^{-\frac{3}{2}}(NI_N - 1_N 1_N^T)$ is the centering matrix and $1_N$ is a column vector of $N$ ones [2,13].

According to Eq.(1), an infinite-dimensional CovDs in RKHS can be defined as:

$$C_{\mathcal{H}} = \varphi(S) J_N J_N^T \varphi(S)^T \quad (2)$$

where $\varphi(S) = [\varphi(s_1), \varphi(s_2) \ldots, \varphi(s_N)]$, $\varphi: R^d \to \mathcal{H}$ is the implicit kernel mapping to $\mathcal{H}$ and the dimensionality of samples in $\mathcal{H}$ approaches $\infty$. Thus, $C_{\mathcal{H}}$ is usually semi-definite and on the boundary of the positive cone.

### B. Affine Invariant Riemannian Metric

The $S_n^+$ is the Riemannian manifold spanned by SPD matrices and can be viewed as a convex cone in the $n(n+1)/2$ dimensional Euclidean space [2]. The similarity between two SPD matrices can be obtained by computing the length of geodesic curve on the SPD manifold, which is analogous to computing the length of straight line between two points in the vector space. The AIRM [2,8] is one of the most popular Riemannian metrics which analyzes the points on the SPD manifold. For a Riemannian point $P$ on the SPD manifold, The AIRM can be defined through two associated tangent vectors $u, v \in T_P S_n^+$:

$$< u, v >_P \triangleq < P^{-\frac{1}{2}} u P^{-\frac{1}{2}}, P^{-\frac{1}{2}} v P^{-\frac{1}{2}} > = tr(P^{-1} u P^{-1} v) \quad (3)$$

According to Eq.(3), the geodesic distance $d_{AIMR}$ between two SPD matrices $Sp_i$ and $Sp_j$ via AIRM [2,8] can be written as:

$$d_{AIRM}(Sp_i, Sp_j) = \left\|\log\left(Sp_i^{-\frac{1}{2}} Sp_j Sp_i^{-\frac{1}{2}}\right)\right\|_F \quad (4)$$

where $\|\cdot\|_F$ is the matrix Frobenius norm, and $\log(\cdot)$ denotes the matrix logarithm operator.

### C. Log-Euclidean Metric

LEM(Log-Euclidean metric) [4,7,8,11] is a bi-variant Riemannian metric, which takes the Riemannian geometry of SPD manifold into account. The distance $d_{\text{LogED}}$ between these two SPD matrices, $Sp_i$ and $Sp_j$, defined via this metric can be written as:

$$d_{\text{LogED}}(Sp_i, Sp_j) = \|\log(Sp_i) - \log(Sp_j)\|_F \quad (5)$$

The meaning of $\|\cdot\|_F$ and $\log(\cdot)$ is the same as that in Eq.(4). LEM can be viewed as the distance between the points in the domain of matrix logarithm, which is the tangent space $S_n$ projected from SPD manifold $S_n^+$ by logarithm mapping [7,8]:

$$\varphi_{\log}: S_n^+ \to S_n, Sp \to \log(Sp), Sp \in S_n^+ \quad (6)$$

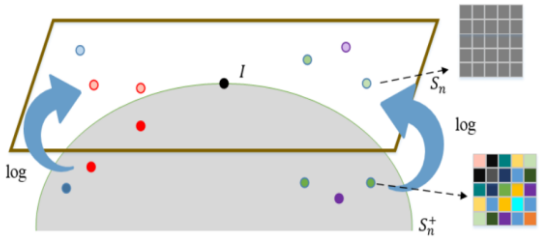

Fig 2. Logarithm mapping

where $S_n$ is a vector space. Figure 2 gives an illustration of the logarithm mapping. Furthermore, the associated Riemannian kernel function can be represented by the inner product of points in the tangent space [7,11]:

$$k_{\text{LogE}}(Sp_i, Sp_j) = < Sp_i, Sp_j >_{\text{LogE}} = tr(\log(Sp_i) \log(Sp_j))$$
$$(7)$$

For all the points $Sp_1, \ldots, Sp_N \in S_n^+$, the final kernel matrix is a symmetric matrix owing to $k_{\text{LogE}}(Sp_i, Sp_j) = k_{\text{LogE}}(Sp_j, Sp_i)$. For any $a_1, \ldots, a_N \in R$, we have:

$$\sum_{i,j} a_i a_j k_{\text{LogE}}(Sp_i, Sp_j) = \sum_{i,j} a_i a_j tr(\log(Sp_i) \log(Sp_j))$$
$$= tr[(\sum_i a_i \log(Sp_i))^2] = \|\sum_i a_i \log(Sp_i)\|_F^2 \geq 0$$
$$a_i \in R, \forall i \in N \quad (8)$$

Eq. (8) proves that the Log-Euclidean kernel guarantees the positive definiteness property of the Riemannian kernel and satisfies the Mercer's theorem.

## III. APPROXIMATE INFINITE-DIMENSIONAL COVARIANCE DESCRIPTORS FOR IMAGE SETS

In this section, we give an overview of the Nyström method which can approximate the SPD matrices in RKHS and introduce our proposed framework to estimate infinite-dimensional CovDs for image sets. We then introduce some classification algorithms for data on SPD manifold which will be used in our experiments.

### A. Nyström Method

As shown in [13] and [20], the Nyström method is a data-dependent estimation method for the vectorial features in RKHS. Here, we extend the Nyström method to approximate SPD matrices in RKHS via Log-Euclidean kernel. Now, given a training set $S = \{Sp_1, Sp_2, \ldots, Sp_M\}$, where $Sp_i \in S_n^+$, consisting of $M$ SPD matrices, then the Riemannian kernel matrix $K = [k(Sp_i, Sp_j)]_{M \times M}$ can be obtained via Eq.(7). The approximation of $K$ can be written as $K \cong Z^T Z = V E^{1/2} E^{1/2} V^T$. Here, $Z = E^{1/2} V^T \in R^{D \times M}$ with $E$ being the diagonal matrix of top $D$ eigenvalues of the Riemannian kernel matrix $K$ and $V$ being the matrix of corresponding eigenvectors. Based on this approximation, a random SPD matrix $Y$ in RKHS can be approximated as a $D$-dimensional vector:

$$Z(Y) = E^{-\frac{1}{2}} V^T (k(Y, Sp_1), k(Y, Sp_2), \ldots, k(Y, Sp_M)) \quad (9)$$

where $Z(Y)$ is the $D$-dimensional vector approximation of the SPD matrix $Y$ in RKHS. For the given training set $S$, it can be estimated as a $D \times M$ matrix $Z(S) = [Z(Sp_1), Z(Sp_2), \ldots, Z(Sp_M)]$, where $Z(Sp_i) \in R^D$ is the $D$-dimensional vector approximation of $Sp_i$ in RKHS. Algorithm 1 summarizes the Nyström method on SPD manifold.

### B. Approximate infinite-dimensional CovDs for image sets

Our proposed framework, which approximates the infinite-dimensional CovDs in RKHS, offers more discriminative and lower-dimensional descriptors for image sets. We firstly

| Algorithm 1: Nyström method on SPD manifold |
|---|
| Input: |
| • training set: $Sp = \{Sp_1, Sp_2, ..., Sp_N\}, Sp_i \in S_n^+$. |
| • $D$, target dimensionality. |
| • a random SPD matrix $Y$. |
| Output: |
| • the approximation of $Y$ in RKHS. |
| 1: compute the Riemannian kernel matrix: $K = [k(Sp_i, Sp_j)]_{N \times N}$ via Eq.(7). |
| 2: obtain the diagonal matrix $E$ of top $D$ eigenvalues of $K$. |
| 3: obtain the corresponding eigenvectors $V$. |
| 4: compute the approximation $Z(Y)$ via Eq.(9). |

| Algorithm 2: Approximate infinite-dimensional CovDs for image sets |
|---|
| Input: |
| • image set: $S = [s_1, s_2 ..., s_N], s_i$ is an image matrix. |
| • $D$, target dimensionality. |
| • training set: $S_{train} = \{s_i\}_{i=1}^M, s_i \in S$. |
| Output: |
| • $C_Z \in S_n^+$, approximate infinite-dimensional CovDs |
| 1: obtain the CovDs of image set $S$ and $S_{train}$: $Sp$ and $Sp_{train}$ via computing covariance matrix of Gabor or SIFT features. |
| 2: obtain associated parameters of Nyström method via Algorithm 1. |
| 3: obtain the approximations $Z(Sp)$ via Eq.(9). |
| 4: compute the approximate infinite-dimensional $C_Z$ via Eq.(11). |

extract the features of the single images via Gabor filter [10,11,13] or SIFT [22] (Fig.1 Step2) and use the CovDs of these features to represent them (Fig.1 Step 3). This offers more robust information for single images, compared to resizing and vectorizing them. Secondly, we map the CovDs into RKHS (Fig.1 Step 4), where the form is implicit and the dimensionality approaches $\infty$. Though the infinite-dimensional CovDs in RKHS offer more discriminative information, its form is implicit and consequently not amenable to extending to infinite-dimensional manifold. To this end, we estimate the infinite-dimensional CovDs in RKHS (Fig.1 Step 5) via an explicit mapping and obtain the finite-dimensional (Fig.1 Step 7) approximations, which offers more discriminative descriptors for image sets.

Consider a set of CovDs, $Sp = \{Sp_1, Sp_2, ..., Sp_N\}, Sp_i \in S_n^+$ corresponding to $N$ images in an image set. According to Eq.(2), the infinite-dimensional CovDs for this image set can be written as:

$$C_{\mathcal{H}} = \varphi(Sp) J_N J_N^T \varphi(Sp)^T \quad (10)$$

where $\varphi(Sp) = [\varphi(Sp_1), \varphi(Sp_2), ..., \varphi(Sp_N)]$, $\varphi$ is a Riemannian kernel mapping. Then, $\varphi(Sp)$ in RKHS can be approximated as: $Z(Sp) = [Z(Sp_1), Z(Sp_2), ..., Z(Sp_N)]$ via Eq.(9). The resulting approximate infinite-dimensional CovDs for image set can be written as:

$$C_Z = Z(Sp) J_N J_N^T Z(Sp)^T \quad (11)$$

where $C_Z$ is the final descriptor extracted for the image set by our proposed framework. The proposed framework is summarized in Algorithm 2.

### C. Classification algorithms based on SPD manifold

The NN(nearest neighbor) algorithm is one of the simplest methods of classification in the domain of computer vision and pattern recognition. In this paper, we use two NN classification algorithms separately based on AIRM and LEM to classify points on the SPD manifold to demonstrate the advantages of our proposed framework for image set classification.

In contrast to [7], where the CDL(covariance discriminative learning) was proposed for image set classification, the classical classification algorithms can be directly utilized on the SPD manifold. In this paper, the Riemannian geometry of the SPD manifold is fully considered. It derives a kernel function that maps the SPD matrices to the Euclidean space through the LEM metric. As a result, the classical classification algorithms applicable in the linear space can be exploited in the kernel formulation. LDA (linear discriminant analysis) and PLS (partial least squares) devoted to the linear space are considered in [7] for the task of classification.

Lastly, we introduce the Riemannian sparse coding algorithm LogEKSR [11], which takes Riemannian geometry of SPD manifold into account and applies the sparse representation in RKHS. In [11], the derivation of the Log-Euclidean kernel is presented to map SPD matrices from SPD manifold to RKHS (Reproducing Kernel Hilbert Space). Note that the Log-Euclidean kernels in this algorithm can be derived from Eq.(7).

Log-EK.poly $\quad k_{p_n}(S, T) = p_n(<S, T>_{LogE})$ (12)
Log-EK.exp $\quad k_{e_n}(S, T) = \exp(p_n(<S, T>_{LogE}))$ (13)
Log-EK.gau $\quad k_g(S, T) = \exp(-||\log(S) - \log(T)||_F^2)$ (14)

where $p_n$ is a polynomial of degree $n \geq 1$ with positive coefficients. These three kernels are respectively the polynomial kernel, exponential kernel, and gaussian kernel associated with LEM. According to [11], these kernels are positive definite and meet the Mercer's theorem.

## IV. EXPERIMENTS AND ANALYSIS

We evaluate the effectiveness of our proposed framework for image set classification. We carry out experiments relating to three different tasks, namely object categorization, hand gesture recognition, and virus cell classification. The three benchmark datasets are ETH-80 [4,21], Cambridge hand gesture dataset (CG) [5] and Virus cell dataset [13] respectively. For benchmarking, we compare the accuracies and standard deviations of traditional CovDs with approximate infinite-dimensional CovDs(proposed) using the same classification algorithm. To this end, we use the nearest neighbor classifier(NN) based on LEM[10] and AIRM[7], which are introduced in Section II. Beside these two NN classifiers, we also make use of another two classification algorithms based on SPD manifold, namely LDA-based CDL (covariance discriminative learning) [7] and LogEK.poly-based LogEKSR

(Log-Euclidean Kernels for Sparse Representation) [11], which are the state-of-the-art method on SPD manifold. The different methods used in our experiments are referred to as:

- NN − AIRM: AIRM-based NN classifier on the SPD manifold spanned by traditional CovDs for image sets.

- NN − AIRM$_{pro}$: AIRM-based NN classifier on the SPD manifold spanned by approximate infinite-dimensional CovDs via our framework for image sets.

- NN − LogED: LEM-based NN classifier on the SPD manifold spanned by traditional CovDs for image sets.

- NN − LogED$_{pro}$: LEM-based NN classifier on the SPD manifold spanned by approximate infinite-dimensional CovDs via our framework for image sets.

- CDL : CDL on the SPD manifold spanned by traditional CovDs for image sets.

- CDL$_{pro}$ : CDL on the SPD manifold spanned by approximate infinite-dimensional CovDs via our framework for image sets.

- LogEKSR : LogEKSR on the SPD manifold spanned by traditional CovDs for image sets.

- LogEKSR$_{pro}$: LogEKSR on the SPD manifold spanned by approximate infinite-dimensional CovDs via our framework for image sets.

To generate the CovDs for single images in image sets, the feature vectors are extracted using Gabor filter [10,11,13] or SIFT [22]. For the Virus cell dataset, the features are extracted using Gabor filter with 5 scales and 8 orientations. For the ETH-80 and CG datasets, the features are extracted using SIFT. For the proposed framework, the dimensionality of the final CovDs is determined by the target dimensionality $D$ in Algorithm 1. We set the target dimensionality $D=40$ and choose 60 training samples at random for the Nyström method on the ETH-80 and Virus datasets, and set the target dimensionality $D=80$ with 104 training samples for the Nyström method on the CG dataset. Thus, the dimensionality of the CovDs obtained by our proposed framework is $80 \times 80$ for the CG dataset and $40 \times 40$ for the ETH-80 and Virus datasets. In the traditional CovDs in our experiments, the dimensionality of the CovDs is $400 \times 400$ via resizing the images to $20 \times 20$.

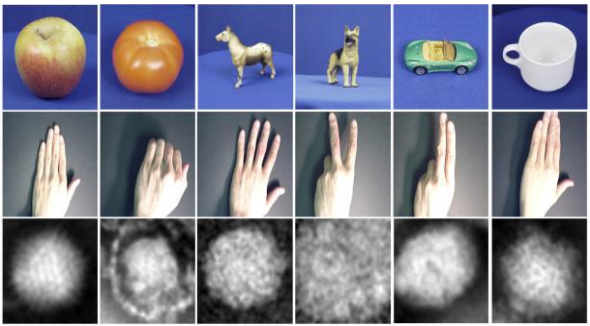

Fig.3: images in three datasets. Top line: ETH-80 [4]. Middle line: CG [5]. Bottom line: Virus [13].

### A. Object Categorization on dataset ETH-80

In the ETH-80 dataset, there are eight categories of images: pears, tomatoes, dogs, cows, apples, cars, horses and cups. Each class consists of 10 image sets, and each set has 41 images from different views (see the top line of Fig.3 for example). For each class, 2 image sets are chosen as training samples randomly and the remaining 8 image sets are used as test samples. The average recognition rate and standard deviation of the 10 cross validation experiments are presented in TABLE I.

### B. Hand Gesture Recognition

For this task, we use the Cambridge hand gesture dataset, which is an image sequence of hand gestures defined by 3 motions and 3 hand shapes. In this dataset, there are 9 categories of images and 900 image sets. Each class has 100 image sets (see the middle line of Fig.3 for example). For each class of samples, 20 image sets are randomly chosen as training samples and the remaining 80 image sets are used as test samples. The average recognition rate and standard deviation of the 10 cross validation experiments are shown in TABLE I.

### C. Virus Vell Classification

The Virus dataset contains 15 categories, each consisting of 5 image sets, and each set has 20 images (see the bottom line of Fig.3 for example). For each class, 3 image sets are randomly chosen as training samples and the remaining 2 image sets are used as test samples. The average recognition rate and standard deviation of the 10 cross validation experiments are used as the final results in TABLE I.

### D. Result and Analysis

TABLE I shows the results of the approximate infinite-dimensional CovDs obtained by our proposed framework, as well as the performance of traditional CovDs using four classification algorithms. On all three datasets, the results achieved by the four classifiers show that the approximate infinite-dimensional CovDs of our proposed framework are more discriminative and robust than the traditional CovDs. Especially on the ETH-80 and Virus datasets, the accuracy of the two NN classifiers, NN-AIRM$_{pro}$ and NN-LogED$_{pro}$, based on our proposed CovDs, is higher than that of the two state-of-the-art classification algorithms, CDL and LogEKSR, based on traditional CovDs. This fully illustrates that our proposed framework offers more discriminative features than the traditional ones. Also, the accuracy of the two classifiers, CDL and LogEKSR, show that our proposed framework is more discriminative. At last, LogEKSR$_{pro}$ achieves the best recognition rates of 89.61% and 63.33%. For the CG dataset, the advantages are not as obvious. However, CDL$_{pro}$ still achieves the best result with accuracy of 91.57% and standard deviation of 0.73.

TABLE I. Results for the ETH-80 [4], CG [5], and Virus [13] datasets.

| Method | ETH-80 [4] | CG [5] | Virus [13] |
|---|---|---|---|
| NN-AIRM | 64.38 ± 5.90 | 54.77 ± 2.36 | 27.00 ± 4.12 |
| NN-AIRM$_{pro}$ | 88.13 ± 4.79 | 69.61 ± 3.76 | 52.44 ± 3.66 |
| NN-LogED | 69.30 ± 5.94 | 70.99 ± 1.59 | 26.30 ± 4.69 |
| NN-LogED$_{pro}$ | 88.36 ± 4.95 | 81.81 ± 4.79 | 54.44 ± 3.67 |
| CDL | 80.14 ± 6.69 | 90.12 ± 1.59 | 46.40 ± 5.76 |
| CDL$_{pro}$ | 89.22 ± 3.12 | **91.57 ± 0.73** | 62.22 ± 5.64 |
| LogEKSR | 86.97 ± 4.48 | 90.20 ± 1.21 | 46.20 ± 6.14 |
| LogEKSR$_{pro}$ | **89.61 ± 2.55** | 91.42 ± 0.36 | **63.33 ± 4.22** |

## V. CONCLUSION AND FUTURE WORK

In this paper, we proposed a novel framework to approximate infinite-dimensional CovDs to extract novel descriptors of image sets. Our experimental results show that the CovDs obtained by our proposed framework are more discriminative than the traditional CovDs in different tasks of image set classification. More importantly, we take the Riemannian geometry of SPD manifold into account because the dimensionality of our proposed CovDs is finite, and our dimensionality is lower than the traditional ones. For the future work, we will consider how to extend our proposed framework to other types of Riemannian manifolds.

## VI. ACKNOWLEDGMENTS

THE PAPER IS SUPPORTED BY THE NATIONAL NATURAL SCIENCE FOUNDATION OF CHINA (GRANT NO.61373055，61672265), UK EPSRC GRANT EP/N007743/1, MURI/EPSRC/DSTL GRANT EP/R018456/1, AND THE 111 PROJECT OF MINISTRY OF EDUCATION OF CHINA (GRANT NO. B12018).